# a self-adaptive system of systems architecture to enable its ad-hoc scalability

Unmanned Vehicle Fleet - Mission Control Center Case study

AHMED R. SADIK, Honda Research Institute Europe, Offenbach am Main 63073, Germany

BRAM BOLDER, Honda Research Institute Europe, Offenbach am Main 63073, Germany

PERO SUBASIC, Honda Research Institute USA, CA 95134, United States

A System of Systems (SoS) comprises Constituent Systems (CSs) that interact to provide unique capabilities beyond any single CS. A key challenge in SoS is ad-hoc scalability, meaning the system size changes during operation by adding or removing CSs. This research focuses on an Unmanned Vehicle Fleet (UVF) as a practical SoS example, addressing uncertainties like mission changes, range extensions, and UV failures. The proposed solution involves a self-adaptive system that dynamically adjusts UVF architecture, allowing the Mission Control Center (MCC) to scale UVF size automatically based on performance criteria or manually by operator decision. A multi-agent environment and rule management engine were implemented to simulate and verify this approach.

**CCS CONCEPTS** • System of Systems • Unmanned Vehicle Fleet • Self-adaptive Architecture

**Additional Keywords and Phrases:** Ad-Hoc Scalability, Holonic architecture, Multi-agent Simulation

## INTRODUCTION

The System of Systems (SoS) terminology was created through multiple evolutionary steps [14]. One of the first notable definitions of SoS was introduced in [2] as "[a] SoS is an array that is a large collection of Constituent Systems (CSs) functioning together to achieve a common purpose." A SoS is evolutionary and emergent. SoS evolution comes as a result that it is composed of CSs that are integrated together to satisfy a higher purpose. Accordingly, SoS are constructed bottom-up. Thus, they are shaped based on an adapting the interconnected CSs to reach a specific purpose. Therefore, the existence of a SoS is evolutionary as tasks are added, removed, or modified during the operation [13]. Ultimately, SoS are emergent [5], which means that SoS behaviors are not fully understood until they are integrated. This implies that the overall SoS value does not equal the sum of the CSs values. Accordingly, some of the overall SoS behaviors can be desirable, while other behaviors might be undesirable.

Swarm robotics is an excellent example of a SoS [19] where multiple robots from the same type that cooperate or operate in parallel to achieve specific tasks that match their capabilities. An Unmanned Vehicle (UV) can be seen as a robot can be remotely controlled by a remote pilot or autonomously controlled by a complex dynamic system based on preprogrammed plans [6]. Different UV categories such as Unmanned Aerial Vehicle (UAV), Unmanned Ground Vehicle (UGV), Unmanned Surface Vehicle (USV), Unmanned Underwater Vehicle (UUA) offer different capabilities, therefore each category can achieve specific missions [1]. There are many advantages of using UVs [8] due to their low cost, high versatility, and easy deployment. However, UVs may also have many limitations such as their mission capacity, mission complexity, operation range, battery lifetime, cybersecurity, and general malfunctioning. Subsequently, using a UV within a fleet becomes the de-facto approach to overcome a single UV limitation [3].

This research defines the Unmanned Vehicle Fleet (UVF) as a SoS that combines many UV constituent swarms, each of which cooperate or collaborate to accomplish complex missions by forming collective structures and emerging behaviors that extend their overall capabilities [18]. The ad-hoc scalability of an UVF is the key to its evolution and emergence. The ad-hoc scalability can be defined as the measure of how responsive and robust the system is to changes that occur during operation by adding or removing resources to meet new demands [9]. Accordingly in this research we provide an approach to its design and simulate a Mission Control Center (MCC) that can control the UVF ad-hoc scalability via dynamically adapting the SoS architecture. In the next section, the research problem is described in detail. Furthermore, section 3 will highlight the central, hierarchical, and holonic system architectures as the fundamentals of the proposed solution. The implementation of the proposed solution will be realized in section 4 via a multi-agent environment that models the MCC, the operator, and the UVs as software agents. Section 5 will discuss the simulation case study and its results. Finally, section 6 will summarize and discuss the research and wrap it up with possible future work.

**PROBLEM**

As has been discussed previously, the UVF can be seen as the realization of the SoS concept over multi-UV Swarms. Swarm robotics is the study of how large number of relatively simple, physically embodied agents can be designed such that a desire collective emerges from the local interaction among the agents and between the agents and the environment [7]. This definition is aligned with the SoS definition in highlighting the scalability as the main key of its emergency and evolution. As shown in Figure 1, a UVF involves two types of ad-hoc scalability. The first type can be seen within a UV swarm (UGV or UAV), where adding/removing UV entities is directly influencing the CS capacity in executing specialized tasks that are suitable with the UV capabilities. The second type can be seen within the overall UVF, where including or excluding an entire UV swarm directly influences the SoS's existing capabilities and overall plan.

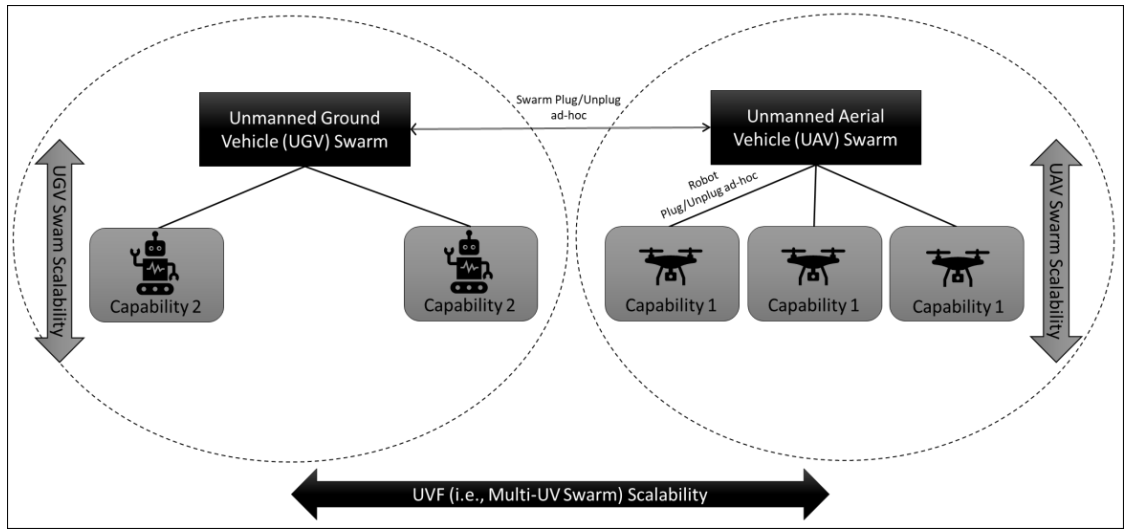

Figure 1: Ad-hoc Scalability in the UVF

The fact that the scalability of a UVF is limited by the static design of its system architecture pattern, that is often implemented by the MCC, is the fundamental challenge that has been highlighted. The main research question that is addressed by this research is "*How to achieve the ad-hoc scalability concept in a UVF by adapting its architecture based on its performance criteria such as battery utilization and communication traffic*"

**SOLUTION**

System architecture patterns are the blueprint to implement an application design at the highest level of abstraction. Architecture patterns defines the overall system structure and behavior. Accordingly, we propose a MCC design that is capable of dynamically scaling the UVF performance by reforming its architecture pattern. Thus, the UVF can easily adapt to unexpected new requirements. The UVF size can be scaled during run time without interruption to its mission. The MCC implements a decision-making mechanism either in fully automatic operation mode or supported-manual operation mode. In fully automatic mode, the MCC autonomously selects the appropriate UVF architecture pattern to fit the current mission requirements. In supported-manual operation mode on the other hand, an operator chooses a specific architecture pattern. Here the MCC validates choices and provides a viable solution. Figure 2 shows the solution concept used to design the MCC. In this concept, the three architecture patterns *Central*, *Hierarchal*, and *Holonic* have been used to demonstrate the solution idea. However, the concept can be applied over other system patterns as well.

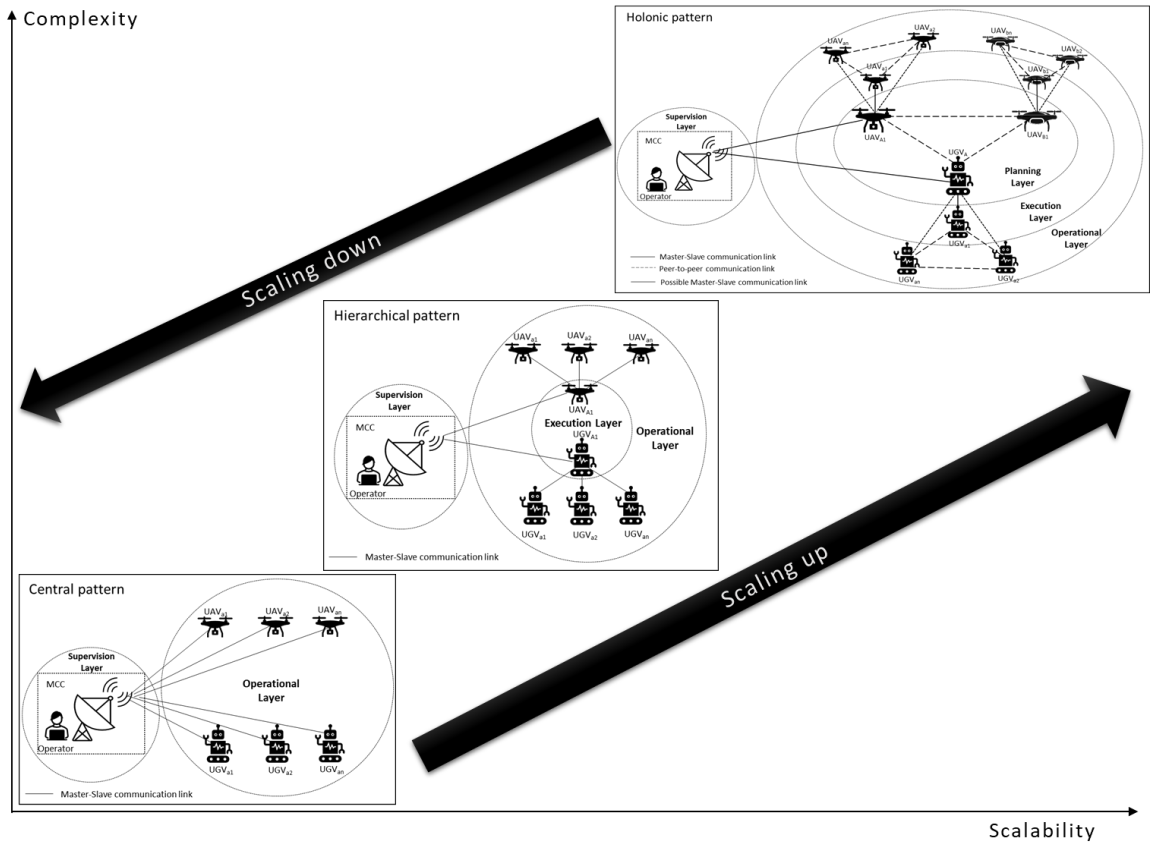

Figure 2: Solution concept – scaling the UVF via adapting its pattern architecture by the MCC.

The *Central* architecture pattern can be seen on the left side of Figure 2. Two structure layers are proposed in this pattern. The first layer is the supervision layer, where the MCC is located. The second layer is the operational layer, where the UVs exist. All the communication links between the MCC and the UVs are based on a master-slave protocol, where in this case the MCC takes the role of the master node while the UVs take the role of slave nodes. In the master-slave protocol, one device acts as a master node that manages all timing and data flow. The slave nodes cannot initiate any communication or communicate with one another. All UVs must therefore wait for the MCC command and then provide their feedback. Would the UVs like to cooperate, they must communicate through the MCC as a hub. The *Central* pattern has a simple structure and behavior, however it incorporates a Single Point of Failure (SPoF) which reduces its reliability [13]. The scalability and flexibility of the *Central* architecture is very limited due to the number of UVs that can be connected and controlled by one MCC simultaneously.

The *Hierarchical* architecture pattern can be seen on the center of Figure 2. It is a modification of the *Central* pattern in a way that increases its scalability and flexibility. Three structural layers are proposed in this pattern. All UVs in the execution layer receive their direct commands from the MCC (located in the supervision layer) via normal master-slave communication links. The UVs in the execution layer are leader UVs, as each of them lead a group of a follower UVs in the operational layer via master-slave communication links. Global cooperation must occur via the MCC and the leader UVs, which naturally increases delays. While the local cooperation can occur via the leader UVs. The *Hierarchical* pattern provides responsive and more reliable performance than the *Central* one, as multiple PoFs are distributed in many locations within the architecture [4]. The UVF can thus partially operate in case of one of the leader UV's failure. However, the follower UVs are helpless without their leaders. Thus, it remains hard to achieve fault-tolerance, which negatively influences the system robustness. Furthermore, the system scalability can be only achieved at the bottom of the hierarchy.

The *Holonic* architecture pattern can be seen in upper right part of Figure 2. Four structural layers are proposed in this pattern. All the UVs in the planning layer receive their direct commands from the MCC in the supervision layer via master-slave communication links. All UVs in this layer can communicate with each other via peer-to-peer communication links. In a peer-to-peer protocol all nodes behave equally. Any node can initiate communication when events occur. Data is exchanged either with one other peer or broadcasted to all the UVs. Moreover, these UVs have a permanent master-slave communication link with one responsible UV in the execution layer. The UV in the execution layer is responsible for executing the plan along with the other peer UVs in the operational layers. In case a UV in the execution layer fails, the UV in the planning layer can promote one of the UVs in the operational layer into the execution layer. It thus establishes a permanent master-slave link with this new UV in the execution layer. The scalability and flexibility of the holonic pattern is extremely high as it can integrate enormous number of UVs. Furthermore, the ability of executing complex plans that involve different capabilities is achievable. Last but not least, this architecture pattern is very robust as an SPoF does practically not exist [17].

**IMPLEMENTATION**

There is no doubt of a large similarity between Multi-Agent Systems (MAS) and the UVF. On the one hand, a MAS is a group of loosely interacting artificial agents, teaming together in a flexible distributed topology to solve a problem beyond the capabilities of a single agent [12]. On the other hand, a UVF is a set of UV Swarms (i.e., CSs) that interact to provide an emerging behavior that none of the CSs can provide on its own [11]. According to this analogy, the implementation of the proposed solution has been based on the MAS environment inside Java Agent DEvelopment (JADE), as shown in Figure 3-a.

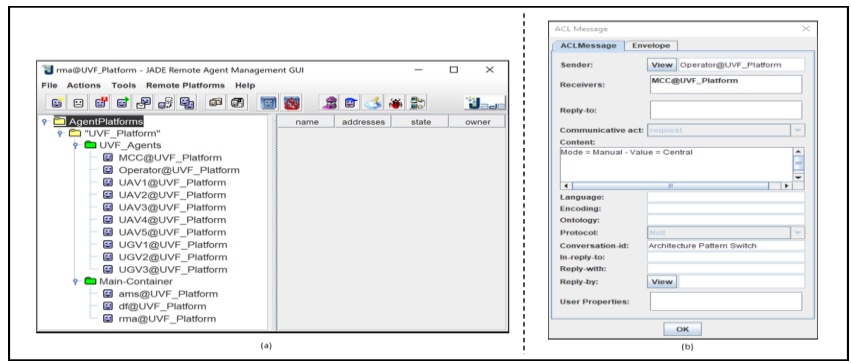

Figure 3: (a) UVF implement as a MAS in JADE (b) ACL message communication

Figure 3-a shows the JADE implementation of the UVF case study under UVF platform. The UVF platform is composed of two containers for software agents. The main container contains the two essential agents – the Agent Management System (AMS) and the Directory Facilitator (DF). The AMS provides a unique ID for every agent to be used

as a communication address, while the DF announces the services which each agent can afford. The UVF-agents container bundles three different agent categories. The first are humans instantiated as the operator agent, the second is the controller which is plays the role of the MCC agent and the third category are the autonomous machines that represent the UVs.

JADE agents use the Foundation for Intelligent Physical Agent - Agent Communication Language (FIPA-ACL) to exchange messages [10]. One example of FIPA-ACL message exchange can be seen in

Figure 3-b. The ACL message that is shown in

Figure 3-b explains how the operator agent can override the control mode from automatic to manual, as the message is sent from the operator agent to the MCC agent. The message content field contains the assigned new value of the required architecture pattern by the operator, which in this case is *Central*. The same message exchange mechanism is followed by the other agents within the UVF platform. The only difference that the operator agent's decision-making and behavior is based on direct interaction with a human operator. However, the decision making in case of the MCC, and the UVs agents is done autonomous, based on a set of rules that are applied by the MCC and the UV behavioral state machine.

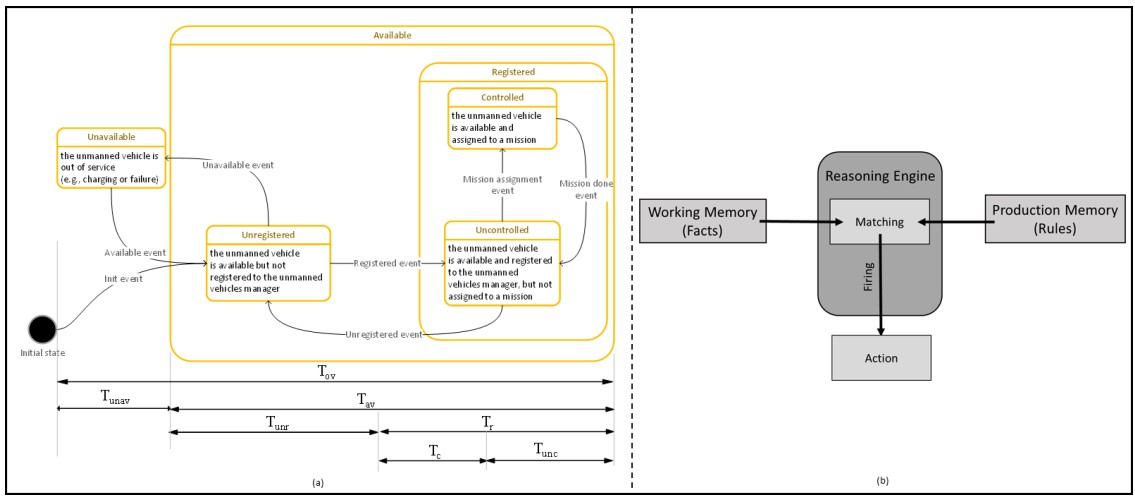

Figure 4: (a) UV behavioral state machine (b) Drools rule engine

Figure 4-a shows the state machine used by a UV agent to model UV behavior. The state machine diagram is a dynamic behavioral diagram described in SysML [16]. It shows the sequences of states that a UV go through during its operation in response to events that may trigger an action. The following UV states are defined:

- Initial: a simple state, where a UV initially becomes ready to operate

- Available: a composite state, where the UV is either registered or unregistered. $T_{av}$ is the time which the UV is available, and either registered or unregistered.
- Unavailable: a simple state, where the UV cannot be registered. It might be out service due or charging its battery. $T_{unav}$ is the time the UV is unavailable.
- Unregistered: a simple state, where the UV is available but not registered yet while it is configuring its parameters. $T_{unr}$ is the time the UV is available but unregistered.
- Registered: a composite state, where the UV is either controlled or uncontrolled. $T_r$ is the time the UV is registered and either controlled or uncontrolled
- Uncontrolled: a simple state, where the UV is registered but not assigned to any mission yet. $T_{unc}$ is the time the UV is registered and uncontrolled
- Controlled: a simple state, where the UV is registered and assigned to a mission. $T_c$ is the time the UV is registered and controlled

Initially, a UV becomes available and unregistered after it receives an init event. If the UV succeeded to configure itself, it will send a registration event to UV's manager. When the MCC accepts, it will be registered but not controlled until the MCC assigns a mission. Then the UV is in the controlled state. When the UV finishes its mission, it goes back into the uncontrolled state. If the UV needs to be reconfigured, it must return to the unregistered state. If it encounters a failure or requires a recharging of the battery, it must return to the unavailable state. Based on this state machine diagram, the MCC can monitor the UVs' statuses and can calculate their Utilization. The UV utilization ($U_{uv}$) is the ratio between $T_c$ to $T_{unc}$. Furthermore, the MCC calculates the UV communication traffic ($Tr_{uv}$), which is the data volume communicated through the UV. $U_{uv}$ and $Tr_{uv}$ are two important performance criteria that are used by the MCC to balance the UVF usage while adapting it is architecture pattern, as it will be illustrated in detail in the next section.

Drools is a deliberative software agent that codifies knowledge bases and reasoning into facts, rules, and actions [15]. Figure 4-b shows the Drools rule engine that has been implemented for the MCC agent to contain the operation rule. The working memory holds the facts presenting the domain knowledge, while the production memory contains the rules represented in form of conditional statements. The reasoning engine is a problem solver that solves a given problem by matching the present facts with the existing rules. The Drools reasoning engine can apply a hybrid chaining reasoning. A hybrid chaining reasoning is a mix between the forward and the backward chaining which can be more efficient in some cases than both.

**CASE STUDY**

The test cases have been deployed in JADE during a 20 min simulation scenario. In this simulation scenario a group of UAVs and UGVs are randomly changing their state based on a set of constrains. The MCC applies either automatic or manual control upon the operator choice. Eventually, the MCC adapts the existing UVs to one of the previously described architecture patterns based on the UVs utilization and traffic that is continuously monitored.

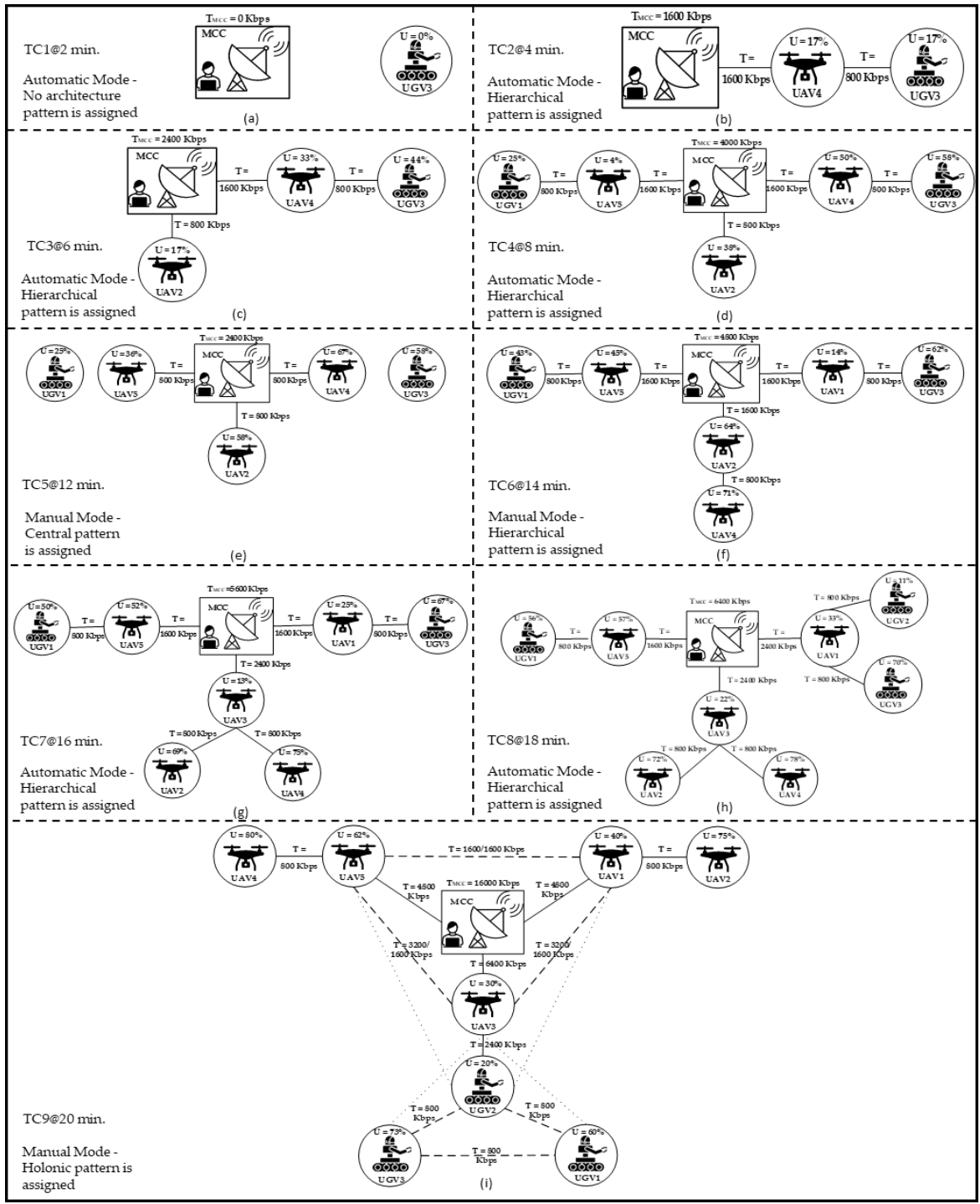

Figure 5: different architectural structure based on the test cases

**Constraints**

- **C1**: the maximum number of available UAVs is five.
- **C2**: the maximum number of available UGVs is three.
- **C3**: all the available UAVs are within the MCC control range.
- **C4**: all the available UGVs are out of the MCC control range.
- **C5**: all the UAVs have similar capabilities, thus any of them can operate as a leader UAV in a cluster.
- **C6**: all the UGVs have similar capabilities, thus any of them can operate as a leader UGV in a cluster.
- **C7**: each of the connected UVs transmits fixed data rate of 800 Kbit.
- **C8**: Maximum number of communication links between the MCC and the UVs is three.
- **C9**: Maximum number of communication links between a leader UV and the follower UVs is two.

**Rules**

- **R1**: in central pattern, one layer (i.e., operational) of UVs that all in the MCC control range can exist.
- **R2**: in hierarchical pattern, two layers (i.e., operational, and execution) of UVs can exist.
- **R3**: in holonic pattern, three layers (i.e., operational, execution, and planning) of UVs can exist.
- **R4**: a UV out of the MCC range is connected either in the execution layer or in the planning layer, via a UV in the operational layer.
- **R5**: In automatic operational mode, the UVs are filling the operational layer at first, then the execution layer, then the planning layer.
- **R6**: the UVs with minimum utilization have higher priority to be directly connected to the MCC (i.e., operational layer must be formed from the UVs with minimum utilization).
- **R7**: connecting a follower UV to a leader UV is based on balancing the traffic on the current leader UVs.
- **R8**: connecting a follower UV to a leader UV is based on balancing the utilization of the current leader UVs, if all the leader UVs have the same traffic.
- **R9**: the third layer of the holonic pattern is composed of clusters from the same UV types.

**Test cases**

- **TC1**: at simulation time of 2 min, the operation mode is automatic. In Figure 5-a, $UGV_3$ is registered but cannot be controlled as it is out of the MCC control range. No architecture pattern is assigned as no UVs are controlled.
- **TC2**: at simulation time of 4 min, the operation mode is automatic. In Figure 5-b, $UAV_4$ is registered and can be controlled. Therefore, $UAV_4$ establishes a communication link with the MCC. Furthermore, as R4 is applied, $UGV_3$ can be connected to the MCC through $UAV_4$ as well. As two communication layers exist, the MCC structures the UVs in hierarchal pattern, according to R2.
- **TC3**: at simulation time of 6 min, the operation mode is automatic. In Figure 5-c, $UAV_2$ is registered. Therefore, the MCC applies R5, to establish a communication link with $UAV_2$, and keep the hierarchal pattern, according to R2.
- **TC4**: at simulation time of 8 min, the operation mode is automatic. In Figure 5-d, $UAV_5$ and $UGV_1$ are registered. Therefore, the MCC applies R5, to establish a communication link with $UAV_5$. Then, the MCC applies R4 to connect $UGV_1$ through $UAV_5$. As two layers of communication still exist, the MCC structures the exiting UV in hierarchal pattern, according to R2.

- **TC5**: at simulation time of 12 min, the operation mode is switched by the operator to manual mode, the assigned pattern is centered. In Figure 5-e, the MCC applies R1 to achieve the central pattern structure.
- **TC6**: at simulation time of 14 min, the operation mode is switched by the operator to manual mode, the assigned pattern is hierarchal. In Figure 5-f, $UAV_1$ has newly registered to the MCC, while $UGV_1$ and $UGV_3$ were already registered but not controlled in TC5. Accordingly, the MCC applies R2, R4, R5, R6, and R7 to achieve the hierarchal pattern structure. According to R5 and R6, the connections to the MCC (i.e., the operational layer) must be completed first, with the UVs with minimum utilization. $UAV_1$, $UAV_2$, $UAV_4$, and $UAV_5$ can directly be connect to the MCC. Based on Figure 6, $U_{A1}$, $U_{A2}$, $U_{A4}$, and $U_{A5}$ are 0%, 58%, 67%, and 36% respectively. Therefore, Based on C8, $UAV_1$, $UAV_2$, and $UAV_5$ form the operational layer. Furthermore, the MCC applies R2, R4, and R7 to form the execution layer. As the MCC applies R4, it finds out that it is supposed to connect three leaders UVs (i.e., $UAV_1$, $UAV_2$, and $UAV_5$) to three follower UVs (i.e., $UAV_4$, $UGV_1$, and $UGV_4$). This means that each leader UV is connected to only one follower UV, to balance the leader UVs balance. The selection of the follower UV therefore occurs randomly, as all the possibilities achieve the MCC rules.
- **TC7**: at simulation time of 16 min, the operation mode is switched by the operator to automatic mode. In Figure 5-g, $UAV_3$ has newly registered to the MCC. Therefore, the MCC applies R6 to minimize the utilization of the operational layer. Based on Figure 6, $U_{A1}$, $U_{A2}$, $U_{A3}$, $U_{A4}$, and $U_{A5}$ are 14%, 64%, 0%, 71%, and 45% respectively. Therefore, Therefore, $UAV_1$, $UAV_3$, and $UAV_5$ form the first UV layer. Furthermore, the MCC applies R2, R4, R7, and R8 to form the execution layer. As the MCC applies R7, it finds out based on Table 1 that the traffic on the leader UVs ($UAV_1$, $UAV_3$, and $UAV_5$) cannot balanced, as one UV must handle 1600 Kbit. Accordingly, the MCC uses R8 to find the leader UV with minimum utilization. Therefore, the MCC selects $UAV_3$ ($U_{A3} = 0\%$), to handle the highest traffic in the execution layer (i.e., 1600 Kbit). As the MCC connects all the registered UVs in a two-layer structure, it concludes from R2 that the current pattern is hierarchal.
- **TC8**: at simulation time of 18 min, the operation mode is still automatic. In Figure 5-h, $UGV_2$ has newly registered. According to R7, $UGV_2$ cannot be led by $UAV_3$. Because $Tr_{A3}$ is 1600 Kbit as shown in Table 1. Therefore, the MCC applies R8 to select $UAV_1$ or $UAV_5$ to lead $UGV_2$. In Figure 6, $U_{A1}$, and $U_{A5}$ are 25% and 52% respectively. Thus, $UAV_1$ is selected to lead $UGV_2$. As the MCC connects all the registered UVs in a two-layer structure, it concludes from R2 that the current pattern is hierarchal.
- **TC9**: at simulation time of 20 min, the operation mode is switched by the operator to manual mode, the assigned pattern is holonic. In Figure 5-i, the MCC applies R5 and R6 to form the operational layer. Therefore, the MCC connects the UVs with minimum utilization. Based on Figure 6, $U_{A1}$, $U_{A2}$, $U_{A3}$, $U_{A4}$, and $U_{A5}$ are 33%, 72%, 22%, 78%, and 57% respectively. Therefore, Based on C8, $UAV_1$, $UAV_3$, and $UAV_5$ form the operational layer. Based on R9, $UGV_1$, $UGV_2$, and $UGV_3$ can form a cluster. As it is expected that the traffic from connecting this cluster is higher than the traffic from connecting $UAV_4$ and $UAV_2$. The cluster is connected to the UAV with minimum utilization within the operational layer, which is $UAV_3$ ($U_{A3} = 22\%$) in this case. As $U_{G1}$, $U_{G2}$, and $U_{G3}$ are 56%, 11%, and 70% respectively. Therefore, based on R8 and C9, $UGV_2$ will have a master-slave link with $UAV_3$. Simultaneously, $UGV_2$ will have a peer-to-peer links with $UGV_1$ and $UGV_3$. According to R7, the MCC concludes that $UAV_1$ can lead only one follower UV, and the same for $UAV_5$, to balance the traffic among them. Accordingly, The MCC assign $UAV_2$ to be led by $UAV_1$, and $UAV_4$ to be led by $UAV_5$.

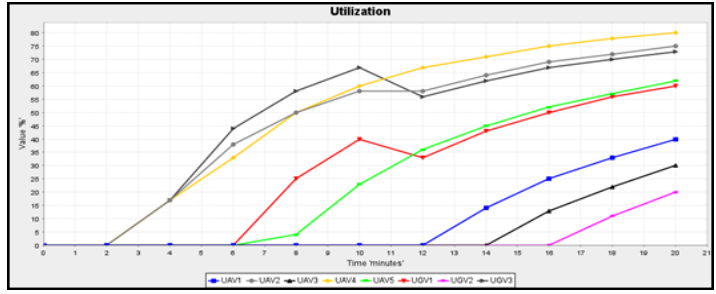

Figure 6: UVs Utilization in min

Table 1: UVs Traffic in Kbit

| Test case | Time (min.) | $Tr_{A1}$ (Kbit) | $Tr_{A2}$ (Kbit) | $Tr_{A3}$ (Kbit) | $Tr_{A4}$ (Kbit) | $Tr_{A5}$ (Kbit) | $Tr_{G1}$ (Kbit) | $Tr_{G2}$ (Kbit) | $Tr_{G3}$ (Kbit) | $Tr_{MCC}$ (Kbit) |
|---|---|---|---|---|---|---|---|---|---|---|
| 1 | 2 | 0 | 0 | 0 | 0 | 0 | 0 | 0 | 0 | 0 |
| 2 | 4 | 0 | 0 | 0 | 800 | 0 | 0 | 0 | 0 | 1600 |
| 3 | 6 | 0 | 0 | 0 | 800 | 0 | 0 | 0 | 0 | 2400 |
| 4 | 8 | 0 | 0 | 0 | 800 | 800 | 0 | 0 | 0 | 4000 |
| 5 | 12 | 0 | 0 | 0 | 0 | 0 | 0 | 0 | 0 | 2400 |
| 6 | 14 | 800 | 800 | 0 | 0 | 800 | 0 | 0 | 0 | 4800 |
| 7 | 16 | 800 | 0 | 1600 | 0 | 800 | 0 | 0 | 0 | 5600 |
| 8 | 18 | 800 | 0 | 1600 | 0 | 800 | 0 | 0 | 0 | 6400 |
| 9 | 20 | 4000 | 0 | 5600 | 0 | 4000 | 1600 | 1600 | 1600 | 16000 |

## DISCUSSION

The research highlighted the challenge of the ad-hoc scalability within a UVF from the SoS prospective. As the main ad-hoc scalability limitation is a result of the static design of the UVF system architecture, this research proposes a dynamic system architecture that adapts its pattern (central, hierarchal, or holonic) based on the UVF demand. A multi-agent simulation has been introduced based on an analogy between the MAS and the UVF characteristics, under the SoS umbrella. The MAS simulation defined three software agent categories, the UV agent, operator agent, and MCC agent.

The UV agent models its autonomous behavior by executing its state machine. This technique provides a realistic simulation leverage, as it enables simulating randomness in the UVs states. For example, during TC1, $UGV_5$ was available. However, upon C4 and R4, $UGV_5$ needed another UV that in the MCC range to be able to communicate with the MCC, which has been fulfilled in TC2. The operator agent enables the human decision making that dramatically influences the flow of events among the other agents. This has been illustrated in TC5, TC6, and TC9, when the operator commands the MCC to assign a specific architecture pattern. The operator decision making enables the flexibility and intelligence that cannot be accomplished by algorithms when dealing with uncertain situation such as cyber-attacks or partial system failure, as they are not designed to resolve these issues. The MCC agent has been implementation using the Drools rule engine. Drools is particularly advantageous in applying forward and backward reasoning simultaneously (i.e., hybrid reasoning). Hybrid reasoning is especially important in automatic operation mode, as the MCC applies the forward reasoning to construct an architecture based on the existing rules and constraints. After constructing the architecture, it uses the backward reasoning to conclude which architecture pattern is assigned. For example, in TC8, the MCC concludes from R2 that the current pattern is hierarchical.

The test cases covered all the rules and constraints. Ultimately, Figure 6 shows that the UVs overall utilization is converging, as the rules tend to balance it over time. Furthermore, Table 1 shows that the traffic is dramatically increasing when switching between TC8 (hierarchal) and TC9 (holonic), due to the increasing of the architecture layers and the peer-to-peer communication. Switching to the central pattern results in zero traffic, as shown in TC5. This means that the cost of a scalable, cooperative, and reliable UVF is reflected on the overall traffic, utilization, and ultimately the UVs battery lifetime. This explains why the existence of the operator is very crucial to comprise all the UVF aspects and cope with uncertain situations. During future work, UVs utilization and traffic will be factorized and expressed as a function of their battery life. The rules can thus use both the utilization and the traffic simultaneously to optimize the UVF architecture pattern.

**REFERENCES**

[1] Godwin Asaamoning, Paulo Mendes, Denis Rosário, and Eduardo Cerqueira. 2021. Drone swarms as networked control systems by integration of networking and computing. Sensors 21. DOI:https://doi.org/10.3390/s21082642

[2] W. Clifton Baldwin and Brian Sauser. 2009. Modeling the characteristics of system of systems. In 2009 IEEE International Conference on System of Systems Engineering, SoSE 2009.

[3] Kathleen Giles. 2016. A Framework for Integrating the Development of Swarm Unmanned Aerial System Doctrine and Design. In Swarm-Centric Solution for Intelligent Sensor Networks, STO-MP-SET-222.

[4] Christian Goerick. 2010. Towards an understanding of hierarchical architectures. IEEE Trans. Auton. Ment. Dev. 3, 1 (2010), 54–63.

[5] Michael Henshaw, Carys Siemieniuch, Murray Sinclair, Vishal Barot, Sharon Henson, Cornelius Ncube, Soo Ling Lim, Huseyin Dogan, Mo Jamshidi, and Dan Delaurentis. 2013. The Systems of Systems Engineering Strategic Research Agenda Systems of Systems Engineering. 8th Int. Conf. Syst. Syst. Eng. Maui, Hawaii, USA - June 2-6, 2013 2 (2013).

[6] Hui Min Huang, Kerry Pavek, Brian Novak, James Albus, and Elena Messina. 2005. A framework for Autonomy Levels for Unmanned Systems (ALFUS). In AUVSI's Unmanned Systems North America 2005 - Proceedings.

[7] Praveen Kalla, Ramji K Ramj, and P Ravindranath. 2017. Swarm home robots. In 2017 IEEE International Conference on Consumer Electronics-Asia (ICCE-Asia), 139–144.

[8] Jinkyu Kim, Teruhisa Misu, Yi-Ting Chen, Ashish Tawari, and John Canny. 2019. Grounding Human-To-Vehicle Advice for Self-Driving Vehicles. In Proceedings of the IEEE/CVF Conference on Computer Vision and Pattern Recognition (CVPR).

[9] Martin Kleppmann. 2019. Designing Data-Intensive Applications.

[10] Sujeet Kumar and Utkarsh Kumar. 2014. Java Agent Development Framework. International Journal of Research 1.

[11] Jérôme Leudet, François Christophe, Tommi Mikkonen, and Tomi Männistö. 2019. AILiveSim: An Extensible Virtual Environment for Training Autonomous Vehicles. In 2019 IEEE 43rd Annual Computer Software and Applications Conference (COMPSAC), 479–488. DOI:https://doi.org/10.1109/COMPSAC.2019.00074

[12] Hengbo Ma, Yaofeng Sun, Jiachen Li, Masayoshi Tomizuka, and Chiho Choi. 2021. Continual Multi-Agent Interaction Behavior Prediction With Conditional Generative Memory. IEEE Robot. Autom. Lett. 6, 4 (2021), 8410–8417. DOI:https://doi.org/10.1109/LRA.2021.3104334

[13] Mark W. Maier. 1998. Architecting principles for systems-of-systems. Syst. Eng. 1, 4 (1998). DOI:https://doi.org/10.1002/(SICI)1520-6858(1998)1:4%3C267::AID-SYS3%3E3.0.CO;2-D

[14] Irene Martin Rubio and Diego Andina. 2018. Smart Manufacturing in a SoSE Perspective. In Advances in Renewable Energies and Power Technologies. DOI:https://doi.org/10.1016/B978-0-12-813185-5.00015-2

[15] Mark Proctor. 2012. Drools: A Rule Engine for Complex Event Processing. In International symposium on applications of graph transformations with industrial relevance, 2–2. DOI:https://doi.org/10.1007/978-3-642-34176-2_2

[16] Ahmed R. Sadik and Christian Goerick. 2021. Multi-Robot System Architecture Design in SysML and BPMN. Adv. Sci. Technol. Eng. Syst. J. 6, 4 (2021). DOI:https://doi.org/10.25046/aj060421

[17] Ahmed R Sadik and Bodo Urban. 2017. Combining adaptive holonic control and ISA-95 architectures to self-orgranize the interaction in a worker-industrial robot cooperative workcell. Future Internet, 9(3), p.35.

[18] Jeffrey Smith, Jayashree Harikumar, and Brian Ruth. 2011. An Army-Centric System of Systems Analysis (SoSA) Definition. (October 2011), 38.